\documentclass{esannV2}
\usepackage[dvipsnames]{xcolor}
\usepackage{graphicx}
\usepackage{amsmath,amssymb}
\usepackage{url,hyperref}
\usepackage{booktabs}  
\usepackage[most]{tcolorbox}

\begin{document}

\title{Cognitive BASIC: An In-Model Interpreted Reasoning Language for LLMs}

\author{
  Oliver Kramer \\
  \small Computational Intelligence Group\\ University of Oldenburg, Germany \\
  \small \texttt{oliver.kramer@uni-oldenburg.de}
}

\maketitle
\thispagestyle{empty}

\begin{abstract}
Cognitive BASIC is a minimal, BASIC-style prompting language and in-model interpreter that structures large language model (LLM) reasoning into explicit, stepwise execution traces. Inspired by the simplicity of retro BASIC, we repurpose numbered lines and simple commands as an interpretable cognitive control layer. Modern LLMs can reliably simulate such short programs, enabling transparent multi-step reasoning inside the model. A natural-language interpreter file specifies command semantics, memory updates, and logging behavior. Our mental-model interpreter extracts declarative and procedural knowledge, detects contradictions, and produces resolutions when necessary. A comparison across three LLMs on a benchmark of knowledge extraction, conflict detection, and reasoning tasks shows that all models can execute Cognitive BASIC programs, with overall strong but not uniform performance.
\end{abstract}

\section{Introduction}

Recent work on \emph{cognitive prompting}~\cite{cp} has shown that LLMs can be guided toward more reliable reasoning when prompts
explicitly reflect cognitive processes such as goal decomposition, declarative
and procedural knowledge extraction, or conflict handling. These approaches move
beyond unstructured text generation by imposing cognitive orientation on the
reasoning steps themselves. However, they still rely on implicit execution: the
model decides how to follow the instructions, and intermediate cognitive states
remain informal and difficult to audit.

Cognitive BASIC  takes the next step in this direction by enforcing structured reasoning
through a minimal in-model programming language. Instead of describing reasoning
procedures at the prompt level, Cognitive BASIC  executes them through a BASIC-style,
line-numbered program interpreted entirely by the LLM. An interpreter file,
written in natural language, defines the semantics of each command, the memory
manipulation rules, and the logging behavior. Programs operate on a compact
working memory containing declarative knowledge (what is known), procedural
knowledge (how to act or reason), detected contradictions, and reconciled
resolutions. Each instruction updates this memory state explicitly, producing a
transparent, auditable reasoning trace.

This design connects two traditions: the transparency aims of cognitive
prompting, and the explicit control flow of early programming languages such as
BASIC~\cite{Kemeny1964}. While prior prompting frameworks, such as
Chain-of-Thought~\cite{Wei2022}, ReAct~\cite{Yao2022}, or modular cognitive
prompts~\cite{cp}, encourage structured steps, they lack an executable semantics.
Classical cognitive architectures including ACT-R~\cite{Anderson2004} and
SOAR~\cite{Laird2012} separate declarative and procedural memory under symbolic
control, and recent agentic systems such as OpenCog Hyperon~\cite{Goertzel2023}
or MemGPT~\cite{Packer2023} offer persistent memory for extended reasoning. Yet
these approaches rely on external engines or customized environments.

\section{Cognitive BASIC  Language and Interpreter}
\label{sec:language}

Cognitive BASIC  adopts the simplicity of early BASIC to structure reasoning inside a language model.  
Programs consist of short, numbered lines executed sequentially unless redirected by control flow.  
The interpreter, defined entirely in natural language, runs within the model and updates a compact memory state after each instruction.

\subsection{Execution Model}
The interpreter follows deterministic BASIC-style semantics~\cite{Kemeny1964}.  
Lines execute in ascending order, with conditional branching through \texttt{IF … THEN <line>} or direct jumps using \texttt{GOTO <line>}.  
After each command, the interpreter applies the operation to the current memory, prints a concise log entry, and proceeds to the next instruction.  
Execution terminates on \texttt{END}, producing a final structured memory state that summarizes all reasoning steps.

\subsection{Memory Schema}

Cognitive BASIC  maintains a compact  memory structure that serves as the
model’s internal mental model during program execution. The variable
\texttt{working} stores the current scenario text or intermediate content and
acts as a short-term buffer for each instruction. The fields
\texttt{declarative} and \texttt{procedural} represent the two central forms of
cognitive knowledge: factual propositions describing what is true, and
operational rules describing how to act or reason. Together, they provide the
basic components of a structured mental model.

Contradictions discovered during execution are recorded in \texttt{conflicts} as
simple string pairs of the form ``A\,\textbar\textbar\,B'', making cognitive
inconsistencies explicit rather than implicit. When a conflict is repaired, the
resulting reconciled statement is stored in \texttt{resolution}, documenting how
the mental model was updated.

\subsection{Instruction Set}
Cognitive BASIC  provides a small but expressive set of BASIC-style commands that operate entirely within the model to control reasoning and memory updates.  
Each instruction interacts with one of the five memory variables, \texttt{working}, \texttt{declarative}, \texttt{procedural}, \texttt{conflicts}, and \texttt{resolution}.  
The syntax mirrors early BASIC conventions: uppercase keywords, lowercase variables, and sequential line-by-line execution unless redirected by control flow.  
This design turns text generation into a transparent sequence of cognitive operations.

\begin{flushleft}\small
\setlength{\tabcolsep}{3pt}
\renewcommand{\arraystretch}{1.05}
\begin{tabular}{@{}lp{0.5\linewidth}@{}}
\toprule
\textbf{Command} & \textbf{Effect} \\
\midrule
\texttt{LET working = INPUT()} &
Load the scenario text into \texttt{working}. \\

\texttt{facts = EXTRACT\_DECLARATIVE(working)} &
Extract declarative facts. \\

\texttt{rules = EXTRACT\_PROCEDURAL(working)} &
Extract procedural knowledge. \\

\texttt{ADD declarative FROM facts} &
Append extracted facts to \texttt{declarative}. \\
\texttt{ADD procedural FROM rules} &
Append extracted procedural steps to \texttt{procedural}. \\

\texttt{DETECT\_CONFLICTS()} &
Identify contradictions between stored facts and populate
\texttt{conflicts} with pairs ``A\,\textbar\textbar\,B''. \\

\texttt{CONFLICTS\_COUNT()} &
Return the number of detected contradictions. \\

\texttt{resolution = RESOLVE\_CONFLICTS()} &
Resolve inconsistencies by generating reconciled statements,
updating \texttt{declarative}, clearing \texttt{conflicts},
and writing a short summary to \texttt{resolution}. \\

\texttt{PRINT <expr>} &
Output a variable or expression to the reasoning log (no state change). \\

\texttt{REM <text>} &
Insert a human-readable comment; ignored by the interpreter (no state change). \\

\texttt{IF CONFLICTS\_COUNT() > 0 THEN <line>} &
Conditionally jump to a specified line if contradictions are present. \\

\texttt{GOTO <line>} &
Unconditionally jump to another line. \\

\texttt{END} &
Terminate execution and print the final memory state. \\
\bottomrule
\end{tabular}
\end{flushleft}

The command \texttt{EXTRACT\_DECLARATIVE()} extracts statements of declarative knowledge, while \texttt{EXTRACT\_PROCEDURAL()} identifies procedural rules or action sequences.
\texttt{DETECT\_CONFLICTS()} checks for inconsistencies among declarative statements, including (i) absolute versus qualified claims such as “always” versus “sometimes” or “never”, (ii) direct negations like “sky is clear” versus “sky is not clear”, and (iii) numeric or categorical disagreements such as “opens at 9” versus “opens at 10”.  
When conflicts arise, \texttt{RESOLVE\_CONFLICTS()} merges opposing claims into qualified summaries (e.g., “usually true but sometimes false” or “uncertain between 9am and 10am”) and records the resolution in \texttt{resolution}.
Together, these mechanisms enable Cognitive BASIC  to model reasoning, contradiction detection, and belief revision as explicit, auditable state transitions inside the LLM.

\subsection{Logging and Output}

During execution, the interpreter produces a transparent, line-by-line reasoning
trace. Each instruction is logged together with a short rationale and the
updated memory state, showing the current contents of the \texttt{working},
\texttt{declarative}, \texttt{procedural}, \texttt{conflicts}, and
\texttt{resolution} fields, as well as the next line to be executed. This
fine-grained trace exposes how the model applies each command, how memory
changes over time, and whether control flow is followed correctly.
At termination, the interpreter prints a final memory block labeled
\texttt{FINAL MEMORY}, containing the complete and internally consistent state
of all variables. 

\section{Experiments and Evaluation}
\label{sec:exp}

Cognitive BASIC  was evaluated on a benchmark of 25 scenarios, each containing contradictory factual statements.  
A single run of the conflict–resolution program therefore jointly tests three cognitive stages:  
(1) extracting propositions into declarative memory,  
(2) detecting their incompatibility as a conflict, and  
(3) producing a coherent reconciled summary that clears the conflict list.  
Preliminary tests on dedicated declarative-only and procedural-only tasks achieved perfect success and are omitted here.

\subsection{Cognitive BASIC  Conflict-Resolution Program}

The full D$\rightarrow$C$\rightarrow$R pipeline is executed inside the model using the following Cognitive BASIC  program, which extracts facts, identifies contradictions, and resolves them when present. Its output is a structured \texttt{FINAL MEMORY} state reflecting each cognitive step.

\begin{tcolorbox}[colback=gray!5!white,
                  colframe=gray!60!black,
                  title=Conflict-Resolution Program,
                  fonttitle=\bfseries\small,
                  left=3mm,right=3mm,top=1mm,bottom=1mm]
\ttfamily\small
10~REM Extract declarative knowledge, detect conflicts, and resolve them\\
20~LET working = INPUT()\\
30~facts = EXTRACT\_DECLARATIVE(working)\\
40~ADD declarative FROM facts\\
50~conflicts\_tmp = DETECT\_CONFLICTS()\\
60~ADD conflicts FROM conflicts\_tmp\\
70~IF CONFLICTS\_COUNT() > 0 THEN 90\\
80~END\\
90~resolution = RESOLVE\_CONFLICTS()\\
100~END
\end{tcolorbox}

\subsection{Evaluation Method}

For each scenario, the model’s execution trace and \texttt{FINAL MEMORY} were examined to determine whether each stage of the pipeline was completed correctly. Declarative extraction was counted as correct if the conflicting statements appeared in the declarative memory. Conflict detection was counted as correct if the conflict list contained a valid contradiction. Conflict resolution was counted as correct if the model produced a coherent reconciled summary and cleared the conflict list. Each scenario yields three binary scores, averaged across all 25 tasks.

Three models were evaluated under identical interpreter and prompting conditions: \texttt{granite3.3}, \texttt{gpt-oss:20b}, and \texttt{mistral:7b}.  
Preliminary trials with smaller models (1B--3B parameters) revealed unreliable program following and incomplete conflict pipelines; these were therefore excluded from the main evaluation.

\subsection{Results and Discussion}

Table~\ref{tab:results} summarizes the performance across the three cognitive
subtasks implemented within the Cognitive BASIC  interpreter: declarative extraction (D),
conflict detection (C), and conflict resolution (R), as well as the complete
D$\rightarrow$C$\rightarrow$R reasoning chain.
All 25 scenarios were processed using the same line-numbered interpreter, ensuring
that differences reflect cognitive reliability rather than prompt variance.

Declarative extraction was solved reliably by all models, confirming that basic
fact parsing is a stable operation under Cognitive BASIC . In contrast, larger
differences emerged in conflict detection and resolution. \texttt{granite3.3}
performed well overall but occasionally failed to recognize numeric or temporal
inconsistencies. \texttt{gpt-oss:20b} showed a stronger degradation: although
declarative extraction remained high, its conflict detection and belief-revision
steps were substantially less reliable, leading to reduced full-chain accuracy.
The smaller \texttt{mistral:7b} model exhibited robust declarative extraction
but showed moderate instability in the conflict pipeline: several contradictions
were missed or resolved incorrectly, yielding an overall full-chain accuracy of~0.80.

\begin{table}[h]
\centering
\small
\setlength{\tabcolsep}{8pt}
\renewcommand{\arraystretch}{1.15}
\caption{Performance of Cognitive BASIC  across declarative extraction (D),
conflict detection (C), conflict resolution (R), and full-chain execution
(D$\rightarrow$C$\rightarrow$R) on 25 scenarios. Scores represent mean accuracy in $[0,1]$.}
\label{tab:results}
\begin{tabular}{lcccc}
\toprule
\textbf{Model} & \textbf{D} & \textbf{C} & \textbf{R} & \textbf{Full Chain} \\
\midrule
granite3.3     & 1.00 & 0.92 & 0.92 & 0.88 \\
gpt-oss:20b    & 0.96 & 0.60 & 0.60 & 0.60 \\
mistral:7b     & 1.00 & 0.84 & 0.80 & 0.80 \\
\bottomrule
\end{tabular}
\end{table}

Taken together, the results indicate that Cognitive BASIC  provides a fine-grained lens on
LLM reasoning stability. Declarative extraction is highly reliable even for small
models, but the multi-step reasoning required for conflict detection and belief
revision remains brittle. Error patterns not only differ across models but also
reveal specific weaknesses: temporal and numeric inconsistencies are harder to
detect, and some models struggle to maintain stable program control flow.  
Cognitive BASIC  thus exposes systematic cognitive failure modes that are difficult to
observe through free-form prompting alone.

\section{Conclusion and Outlook}
\label{sec:conclusion}

Cognitive BASIC  combines a retro programming paradigm with modern in-context learning to
provide an interpretable cognitive control layer that runs entirely inside the
model. Small cognitive programs, such as declarative or procedural extraction and
contradiction handling, execute in a deterministic, line-numbered fashion,
revealing how LLMs manage memory, follow control flow, and revise beliefs.

Future work will extend Cognitive BASIC  with tool-use capabilities that can be invoked
directly during in-model execution. Currently, any operation requiring external
retrieval or computation must be handled by an outside controller before
resuming the program. A more integrated design would allow the model to issue
and incorporate tool calls autonomously. Another direction is hierarchical
control, where each Cognitive BASIC  step is overseen by a higher-level executive agent.
Alternative syntactic designs may also be explored; BASIC was chosen here for
its clarity and its natural fit with stepwise cognitive programs.

\bibliographystyle{unsrt}

\end{document}